\documentclass{article}

\usepackage{PRIMEarxiv}

\usepackage[utf8]{inputenc} 
\usepackage[T1]{fontenc}    
\usepackage{url}            
\usepackage{booktabs}       
\usepackage{amsfonts}       
\usepackage{nicefrac}       
\usepackage{microtype}      
\usepackage{lipsum}
\usepackage{fancyhdr}       
\usepackage{graphicx}       
\graphicspath{{media/}}     

\usepackage{xcolor}         
\usepackage{multirow}
\usepackage{graphicx}
\usepackage{caption}
\usepackage{subcaption}
\usepackage{amssymb}
\usepackage{color}

\usepackage{pifont}
\usepackage{enumitem}

\usepackage{hyperref}
\definecolor{darkblue}{rgb}{0, 0, 0.5}
\hypersetup{hypertex=true, colorlinks=true, citecolor=darkblue, linkcolor=darkblue, urlcolor=darkblue}

\usepackage{marvosym}

\usepackage{wrapfig}

\title{Speak It Out: Solving Symbol-Related Problems with Symbol-to-Language Conversion for Language Models}

\author{Yile Wang$^{1}$ \quad Sijie Cheng$^{1,2,3}$ \quad Zixin Sun$^{2}$ \quad Peng Li\textsuperscript{\Letter}$^{1,4}$ \quad Yang Liu\textsuperscript{\Letter}$^{1,2,4}$ \\
$^1$Institute for AI Industry Research (AIR), Tsinghua University, Beijing, China \\
$^2$Dept. of Comp. Sci. \& Tech., Institute for AI, Tsinghua University, Beijing, China \\
$^3$01.AI \quad $^4$Shanghai Artificial Intelligence Laboratory, Shanghai, China \\
  \texttt{\{wangyile,lipeng\}@air.tsinghua.edu.cn},
  \texttt{liuyang2011@tsinghua.edu.cn}
}

\begin{document}

\maketitle

\begin{abstract}
Symbols (or more broadly, non-natural language textual representations) such as numerical sequences, molecular formulas, and table delimiters widely exist, playing important roles in various tasks such as abstract reasoning, chemical property prediction, and table question answering. 
Despite the impressive natural language comprehension capabilities of large language models (LLMs), their reasoning abilities for symbols remain inadequate, which could attributed to the difference between symbol representations and general natural languages.
We propose symbol-to-language (S2L), a tuning-free method that enables large language models to solve \textit{symbol}-related problems with information expressed in natural \textit{language}.
Specifically, S2L first converts the symbols involved to language-based representations, which can be implemented by prompting LLMs or leveraging external tools, then these language-based representations are integrated into the original problem via direct substitution or concatenation, serving as useful input information for LLMs.
We evaluate the S2L method using both API-based (GPT-4, ChatGPT) and open-source (OpenChat) models over eight symbol-related tasks, ranging from symbol-only abstract reasoning to sentiment analysis in social media. 
Experimental results show that S2L consistently leads to superior performance. For example, by employing S2L for GPT-4, there can be average significant improvements of $+21.9\%$ and $+9.5\%$ for subtasks in 1D-ARC and Dyck language, respectively.
Codes and data are available at \url{https://github.com/THUNLP-MT/symbol2language}.

\end{abstract}
\begin{figure*}[h!]
	\centering
	\includegraphics[scale=1.1]{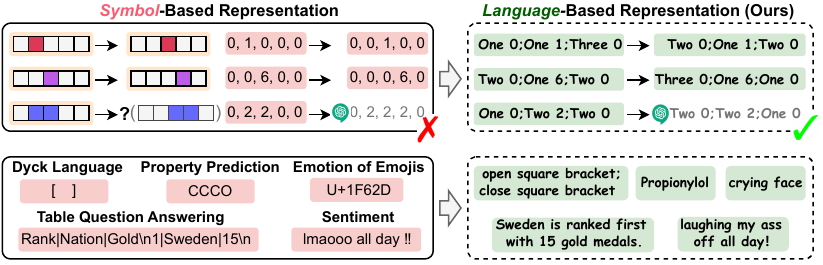}
	\caption{\textbf{Top}: This illustration presents a symbol-related problem (move 1 pixel forward) from the 1D-ARC benchmark, comparing the responses of large language models using both conventional symbol-based and our language-based representations with symbol-to-language (S2L) conversion. \textbf{Bottom}: The S2L conversion has broad applicability across various scenarios involving diverse types of symbols.}
	\label{figure:intro}
\end{figure*}

\newpage

\section{Introduction}\label{sec:intro}

Symbols, or more broadly, non-natural language representations such as brackets, digits, molecular formulas, emojis, table delimiters, and abbreviations are ubiquitously encountered in the real world. 
They are of significant relevance in our daily lives, convey distinct meanings, and play important roles in a variety of tasks.
These symbol-related tasks include abstract reasoning~\citep{moskvichev2023conceptarc,xu2023llms}, chemical property prediction~\citep{ross2022large,guo2023gpt}, and tabular question-answering~\citep{Chen2020TabFact,chen-2023-large}, as exemplified in Table~\ref{table:alltasks}.
Consequently, the understanding and reasoning abilities of symbols are of paramount importance for artificial intelligence~\citep{chollet2019measure}.

Nowadays, large language models (LLMs; \citealp{brown2020language, instructgpt,openaichatgpt,openaigpt4,jiang2023mistral,team2023gemini}) have demonstrated impressive abilities in comprehending and generating natural language. 
GPT-3~\citep{brown2020language} has showcased capabilities of zero-shot inference that solve problems directly without demonstrations. \cite{lets-think-step-by-step} further propose zero-shot-CoT through additional prompting like ``\textit{Let's think step by step}'' to enhance the zero-shot reasoning capability.
However, the understanding and reasoning capability of
symbols for LLMs still fall behind compared to general natural language. For example,~\cite{mitchell2023comparing} reveal that GPT-4~\citep{openaigpt4} and GPT-4V~\citep{2023GPT4V} only achieve an accuracy of 65\% and 25\% on minimal abstract reasoning tasks~\citep{moskvichev2023conceptarc} requiring inductive reasoning through a series of regular numbers or pixels, which is significantly lower than the human accuracy of 95\%.
\cite{gendron2023large} further demonstrate that existing LLMs exhibit limited performance for symbol-related problems in contrast with other natural language tasks.

The inadequate symbol-related reasoning capability of LLMs can be attributed to two primary factors.
First, symbols are significantly underrepresented in the training corpus compared to natural language~\citep{ohsuga2007bridging},
leading to an understanding gap between low-frequency symbols and LLMs~\citep{kandpal2023large, tang2023large}.
Thus, recent studies have explored the collection of symbolic data (\textit{e.g.}, first-order logic, biomolecule, SQL) for continuous training of LLMs~\citep{yang2023harnessing,fang2023mol,xu2023symbol}, which demands massive human annotations and computational resources.
Second, the reasoning capability of LLMs is compromised when dealing with symbolic-related problems due to the subpar understanding of symbol-based representations.
Previous studies~\citep{xu2023symbol,gendron2023large,wang2023hypothesis} directly use these symbol-based representations as inputs for reasoning, while language-based rationales~\citep{wang2022rationale,lets-think-step-by-step,cot} would further induce error propagation due to suboptimal understanding capability of the involved symbols. 


Considering the gap between symbol-based representations and LLMs, our intuition is converting symbols into corresponding language-based expressions.
This conversion serves as a bridge, providing LLMs with more friendly and comprehensible information.
In this paper, we propose symbol-to-language (S2L), a tuning-free method that leverages LLMs to better solve symbol-related problems. 
The S2L method is both simple and feasible, with its core focus on uncovering language-based expressions that are equivalent or approximate to symbols.
In particular, S2L first converts the symbols involved in the problems into their language-based representations, which can either be implemented by prompting the LLMs themselves or by leveraging external cost-friendly tools, such as rules, translators, dictionaries, \textit{etc}.
Then these language-based representations are integrated into the original questions through direct substitution or concatenation, providing valuable contextual information to assist LLMs in solving symbol-related problems.


We conduct experiments across eight symbol-related problems, encompassing inductive abstract reasoning over numerical sequences, Dyck language involving strings of brackets, chemical property predictions based on molecular formulas, emotion analysis of emojis, question answering on structured tabular data, and stance and sentiment analysis in social media. 
We employ both API-based and open-source LLMs including GPT-4~\citep{openaigpt4}, ChatGPT~\citep{openaichatgpt}, and  OpenChat~\citep{wang2023openchat} to validate the generalization of S2L method.
Experimental results demonstrate that the S2L leads to significant and consistent improvements under zero-shot or zero-shot-CoT settings, ranging from symbol-only reasoning to conventional natural language processing tasks involving symbols.
These results underscore the effectiveness of leveraging language-based representations in better addressing symbol-related problems, thereby expanding the potential applicability of LLMs in a broader range of scenarios.

\section{Related Work}
{\noindent \textbf{Reasoning on Symbol-Related Problems.}} 
Various studies have explored the capabilities of LLMs in symbol-based understanding and reasoning. \citet{wang2023hypothesis} suggest that LLMs can improve the resolution of abstract reasoning tasks based on the strong ability to generate executable codes. \citet{qiu2023phenomenal} assess LLMs on symbol-based inductive reasoning tasks, uncovering a range of counter-intuitive behaviors. \citet{gendron2023large} demonstrate that LLMs possess a constrained capacity for abstract reasoning compared with other natural language tasks. These works indicate that there is still room for improvement in the reasoning ability of LLMs for symbol-related problems. Our method tries to bridge the gap between symbol understanding and language models by replenishing language-based representation, eliciting the LLMs' capability of solving symbol-related problems without affecting their general abilities.




{\noindent \textbf{Reasoning by Chain-of-Thought Prompting.}} 
Chain-of-Thought style prompting~\citep{cot,lets-think-step-by-step,chen2022program,besta2023graph,yao2023tree,zhang2023cumulative} have become integral in augmenting the reasoning capabilities of LLMs.
\cite{cheng2023unsupervised} and \cite{tang2023explain} propose to generate readable explanations for solving commonsense and program translation problems, respectively. \citet{deng2023rephrase} introduce rephrase-and-respond to tackle potentially \textit{ambiguous questions} by using self-rephrased questions, which shares similarities with our symbol-to-language method in \textit{rephrasing}. However, we aim to address \textit{symbol-related problems}, rephrasing symbols into their natural language equivalents to allow LLMs to engage with more accessible language-based information for reasoning.

\noindent\textbf{Reasoning with Symbolic Methods.} There is a series of studies on the integration of symbolic methods to solve general reasoning tasks.
\citet{wei2023symbol} propose symbol-tuning to fine-tune LLMs using substituted labels, aiming to bolster their in-context learning abilities. \citet{hu2023chain} propose chain-of-symbol to solve planning-based tasks. \citet{wang2023meta} introduce meta-reasoning as a means to construct generic symbolic representations for reasoning tasks. \citet{fang2024large} design symbolic module in LLM agent for solving text-based games. \citet{AlphaGeometryTrinh2024} combine LLMs and symbolic engines to solve geometry problems. As a comparison, our work focuses on symbol-related tasks and proposes eliciting the powerful comprehension and reasoning abilities of LLMs at the \textit{language} level to help solve the problems.



\begin{table*}[t!]
	\centering
 \scalebox{0.93}{
	\begin{tabular}{lcll}
	    \toprule
        \textbf{Symbol} & \textbf{Instance} &  \textbf{Task} & \textbf{Example}\\
    	\midrule
        Sequence of Numbers & \texttt{1,0,0} & Abstract Reasoning ($\S$~\ref{sec:arc})&\texttt{1,0,0}\ $\rightarrow$\ \texttt{0,0,1};\ \  \texttt{5,0,0}\ $\rightarrow$\ {?}\\
        \midrule
        String of Brackets & [ ( ) ] &  Dyck Language ($\S$~\ref{sec:dyck})&\texttt{([]}\ $\rightarrow$\ \texttt{)};\ \ \texttt{\{(<>)}\ $\rightarrow$\ \texttt{\}};\ \ \texttt{\{\}[}\ $\rightarrow$\ {?}\\
        \midrule
        Molecular Formula & \texttt{CCCO} &Property Prediction ($\S$~\ref{sec:chem})&\texttt{CCCO}\ (\textit{Toxicity: Yes or No}?)\\
        \midrule
        Emoji (Unicode)& \includegraphics[width=3.2mm, height=3.2mm]{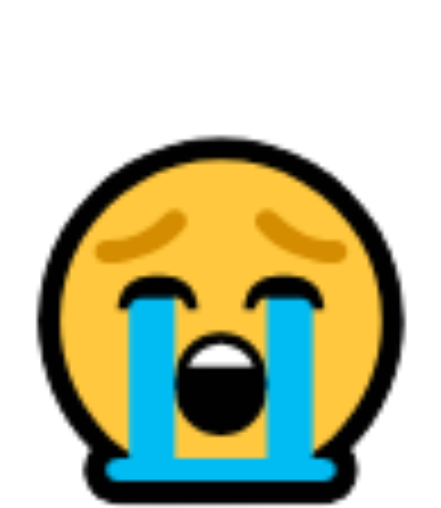} (\texttt{U+1F62D})& Emotion Analysis ($\S$~\ref{sec:emoji})&  \includegraphics[width=3.2mm, height=3.2mm]{figures/emoji9.png} (\textit{Anger}? \textit{Fear}? \textit{Joy}? \textit{Sadness}?)\\
        \midrule
        \multirow{2}*{Table Delimiter} &\multirow{2}{*}{\texttt{|,$\backslash$n}} & Question Answering ($\S$~\ref{sec:table})&\texttt{rank|name|wins$\backslash$n1|jack|3$\backslash$n}\\
        & &Fact Verification ($\S$~\ref{sec:table})&(\textit{Statement: True or False}?)\\
        \midrule
        \multirow{2}*{Abbreviation} & \multirow{2}{*}{\texttt{LMAO}} & Stance Detection ($\S$~\ref{sec:tweet})&\texttt{Imagine being that bold LMAO}\\
        & & Sentiment Classification ($\S$~\ref{sec:tweet})&(\textit{Text: Positive or Negative}?)\\
        \bottomrule
\end{tabular}}

\caption{Symbols, instances, and examples across eight different types of tasks in our experiments, varying from symbol-only inductive abstract reasoning to traditional sentiment classification in social media.}
\label{table:alltasks}
\end{table*}

\section{Symbol-to-Language}
We consider a range of symbol-related problems depicted in Table~\ref{table:alltasks}, in which the interpretation of symbol meanings is crucial for accomplishing the associated tasks.
We formalize this type of problem as a question $q_s$ that incorporates a set of symbols $s = \{s_1,s_2,...,s_m\}$  and try using current LLMs to solve $q_s$.
Vanilla zero-shot~\citep{brown2020language} and zero-shot-CoT~\citep{{lets-think-step-by-step}} method directly solve the original problem and the responses of these two methods by LLM $\mathcal{M}$ can be written as:
\begin{equation}
    {{\mathcal{R}}_{\rm zs}} = {\mathcal{M}} (q_s); \quad {{\mathcal{R}}_{\rm zsc}} = {\mathcal{M}} (q_s \oplus p),
    \label{eq:1}
\end{equation}
where {\rm zs} and {\rm zsc} indicate zero-shot and zero-shot-CoT, respectively. $p$ is a prompt like ``\textit{Let's think step by step}'', and $\oplus$ is the concatenation operation.

The above methods directly tackle the problem with symbol-based representations. Instead of designing external prompt $p$, we focus on the question $q_s$ and propose utilizing language-based representation to better leverage the LLMs' strong capabilities of natural language for solving symbol-related problems. Specifically, the S2L framework first converts the symbols $s_i$ $(i=1,...,m)$ to its corresponding plain text $l_i$ with a conversion operation $f$, which can be implemented by either LLMs themselves or external tools.
Then the S2L framework incorporates the converted language-based representation $l_i$ into two alternative questions $q_{l}$ or $q_{s\oplus l}$ as the input for LLMs to generate answers. The details are discussed as follows.

\subsection{Symbol-to-Language Conversion}
\textbf{Conversion with LLMs.}
We first employ the LLMs $\mathcal{M}$ to convert symbols $s_i$ to their corresponding natural language descriptions $l^{\rm LLM}_i$ via zero-shot prompting, as expressed by:
\begin{equation}
    l^{\rm LLM}_i = f_{\rm LLM} \circ s_i = \mathcal{M} (p_{\rm s2l} \oplus s_i),
    \label{eq:2}
\end{equation}
where $f_{\rm LLM} \circ s_i$ denotes converting $s_i$ using the LLM $\mathcal{M}$, $p_{\rm s2l}$ is the task-specific prompt facilitating the S2L conversion. For instance, when the symbol-related question $q_s$ is about property prediction and $s_i$ is a molecular formula, $p_{\rm s2l}$ could be ``\textit{What does the following molecular formula represent?}''.

\textbf{Conversion with Tools.} 
Considering there exist some constructed ``symbol-language'' pairs, we further propose using external tools for conversion, which can take on several forms.
\textit{Rule-based codes}, for example, can convert $s_i=$ ``\texttt{rank|nation$\backslash$n1|SWE}'' into $l^{\rm rule}_i=$ ``rank: 1; nation: SWE'' according to the table delimiters ``\texttt{|}'' and ``\texttt{$\backslash$n}''.
\textit{Translators} can transform molecular formulas into their formal names, such as converting $s_i=$ ``\texttt{CCCO}'' into $l^{\rm translator}_i=$ ``Propionylo''.
\textit{Unicode dictionaries} can provide descriptions of emojis, like converting $s_i=$ ``\texttt{U+1F62D}'' into $l^{\rm dict}_i=$ ``crying face''.
Despite having some limitations in terms of usage scenarios, conversion with tools offers two primary advantages: 1) it circumvents the costs associated with using LLMs; 2) it provides verified language-based information, which can help reduce potential errors in descriptions generated by LLMs.

\begin{figure*}[t!]
	\centering
	\includegraphics[scale=0.99]{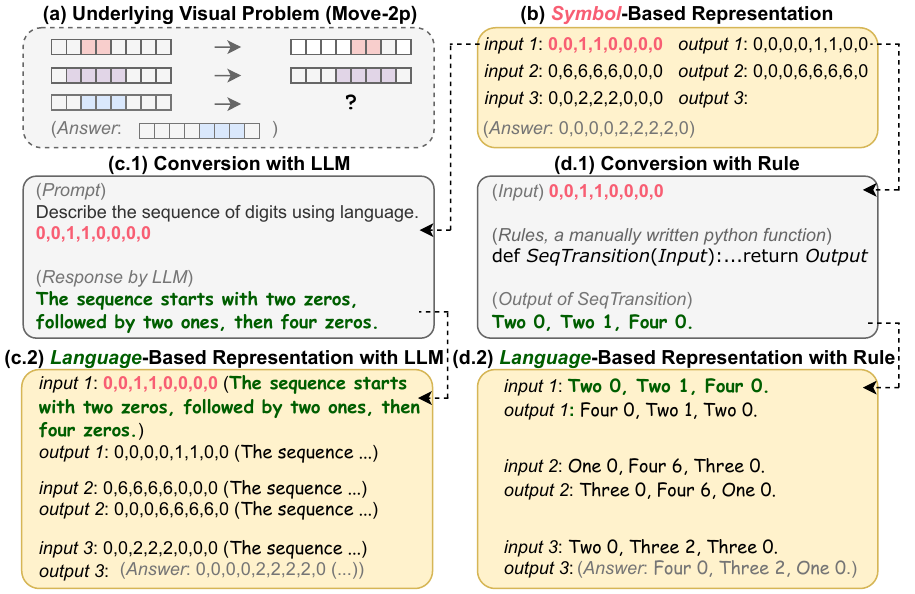}
	\caption{Example of applying symbol-to-language for 1D abstract reasoning task. We convert every sequence to its textual representation by prompting LLMs or using simple rules implemented in code, and then we transform the symbolized problem to language-enhanced or language-only representations.}
	\label{figure:1d-arc}
\end{figure*}

\subsection{Utilizing Language-Based Representations}
We propose two alternative ways that incorporate the language-based representation $l_i$ into the final input. 

\textbf{Direct Substitution.} The first way of utilization is to directly substitute the symbol-based representation $s_i$ with language-based representation $l_i$. To some extent, $l_i$ can be regarded as the language-based equivalent of $s_i$. Thus we can use them to replace the symbol-based representations for both the question and ground-truth label. The response by using S2L can be written as:
\begin{equation}
    {{\mathcal{R}}_{\rm s2l}}= {\mathcal{M}} (q_{l}), \quad 
    q_l = q_{f\circ s} = q_{f\circ \{s_1,...,s_m\}} = q_{f\circ s_1,...,f\circ s_m}.
    \label{eq:3}
\end{equation}

\textbf{Concatenation.}
However, in some other tasks, the generated $l_i$ by LLMs may not always be a perfect substitution or convey complete information of $s_i$. This can occur for two reasons: 1) $l_i$ might be incorrect due to the undesired output formats, misleading content, or noisy context; and 2) $l_i$ could lose some information during S2L conversion.
For instance, the ground truth might be a span-based abbreviation for table understanding (\textit{e.g.}, the ``\texttt{SWE}'' in table ``\texttt{rank|nation$\backslash$n1|SWE}'' ), meaning that the converted $l_i$ as a full name (\textit{e.g.}, ``Sweden'') may not exactly match the final answer.
Therefore, the second method uses both the original symbol-based representation $s_i$ and the language-based representation $l_i$ as the combined input.
This approach allows the LLMs $\mathcal{M}$ to reason on questions that include rich contextual information from two distinct perspectives:
\begin{equation}
    {{\mathcal{R}}_{\rm s2l}}= {\mathcal{M}} (q_{s\oplus l}), \quad q_{s\oplus l} = q_{s_1 \oplus l_1,...,s_m \oplus l_m} = q_{s_1 \oplus \{f\circ s_1\},...,s_m \oplus \{f\circ s_m\}}.
    \label{eq:4}
\end{equation}

\section{Experiments}
\label{sec:experiment}

To assess the performance of the S2L framework, we conduct six categories of symbol-related problems with eight specific tasks as shown in Table~\ref{table:alltasks}, varying from symbol-only inductive abstract reasoning to traditional sentiment analysis in social media. 
As for LLMs, we evaluate both API-based and open-source models, including GPT-4~\citep{openaigpt4}, ChatGPT~\citep{openaichatgpt}, and OpenChat-7b~\citep{wang2023openchat}\footnote{Specifically, we use the versions of \texttt{gpt-4-0613} and \texttt{gpt-3.5-turbo-1106} released by OpenAI, and \texttt{openchat\_3.5} released in \url{https://huggingface.co/openchat/openchat_3.5}.}. To ensure the reproducibility of the responses generated by LLMs, we set the decoding temperature as $0$. 

\begin{table*}[t!]
\scalebox{0.813}{
\begin{tabular}{llllllll}
	    \toprule
        \multirow{2.5}*{\textbf{Model}}&\multicolumn{2}{c}{\textbf{Move-1p}}&\multicolumn{2}{c}{\textbf{Move-2p}}&\multicolumn{2}{c}{\textbf{Move-3p}}&\multicolumn{1}{c}{\multirow{2.5}*{\textsc{\textbf{Avg}.}}}\\
        \cmidrule(lr){2-3}\cmidrule(lr){4-5}\cmidrule(lr){6-7}
        &$n=3$&$n=4$&$n=3$&$n=4$&$n=3$&$n=4$&\\
    	\midrule
     \multicolumn{8}{c}{(\textit{Zero-Shot})}\\
    \texttt{gpt-4}&$93.3$&$90.0$&$48.3$&$46.7$&$35.0$&$45.0$&$59.7$\\
     \ +S2L w/ model&$90.0$~\textcolor{magenta}{\small{($-3.3$)}}&$91.7\phantom{0}$~\textcolor{teal}{\small{($+1.7$)}}&$48.3$~{\small{($-$)}}&$56.7$~\textcolor{teal}{\small{($+10.0$)}}&{$38.3$}~\textcolor{teal}{\small{($+3.3$)}}&{$48.3$}~\textcolor{teal}{\small{($+3.3$)}}&{$\bf{62.2}$}~\textcolor{teal}{\small{($+2.5$)}}\\
     \ +S2L w/ rule&{$96.7$}~\textcolor{teal}{\small{($+3.4$)}}&{$100.0$}~\textcolor{teal}{\small{($+10.0$)}}&{$88.3$}~\textcolor{teal}{\small{($+40.0$)}}&{$96.7$}~\textcolor{teal}{\small{($+50.0$)}}&{$48.3$}~\textcolor{teal}{\small{($+13.3$)}}&{$60.0$}~\textcolor{teal}{\small{($+15.0$)}}&{$\bf{81.6}$}~\textcolor{teal}{\small{($+21.9$)}}\\
     \midrule
     \texttt{gpt-3.5-turbo} &$68.3$&$71.7$&$11.7$&$25.0$&$23.3$&$26.7$&$37.8$\\
     \ +S2L w/ model&{$80.0$}~\textcolor{teal}{\small{($+11.7$)}}&{$75.0$}~\textcolor{teal}{\small{($+3.3$)}}&{$31.6$}~\textcolor{teal}{\small{($+19.9$)}}&{$26.6$}~\textcolor{teal}{\small{($+1.6$)}}&{$33.3$}~\textcolor{teal}{\small{($+10.0$)}}&{$31.7$}~\textcolor{teal}{\small{($+5.0$)}}&{$\bf{46.4}$}~\textcolor{teal}{\small{($+8.6$)}}\\
     \ +S2L w/ rule&{$71.6$}~\textcolor{teal}{\small{($+3.3$)}}&{$88.3$}~\textcolor{teal}{\small{($+16.6$)}}&{$25.0$}~\textcolor{teal}{\small{($+13.3$)}}&{$31.7$}~\textcolor{teal}{\small{($+6.7$)}}&{$25.0$}~\textcolor{teal}{\small{($+1.7$)}}&$26.7$~\small{($-$)}&{$\bf{44.7}$}~\textcolor{teal}{\small{($+6.9$)}}\\
     \midrule
     \texttt{openchat-3.5-7b} & $61.7$ & $71.7$ & $15.0$ & $21.6$ & $11.7$ & $11.7$ & $32.2$ \\
     \ +S2L w/ model & {$68.3$}~\textcolor{teal}{\small{($+6.6$)}} & {$78.3$}~\textcolor{teal}{\small{($+6.6$)}} & {$23.3$}~\textcolor{teal}{\small{($+8.3$)}} & {$25.0$}~\textcolor{teal}{\small{($+3.4$)}} & {$25.0$}~\textcolor{teal}{\small{($+13.3$)}} & {$21.7$}~\textcolor{teal}{\small{($+10.0$)}} & {$\bf{40.3}$}~\textcolor{teal}{\small{($+8.1$)}} \\
     \ +S2L w/ rule & {$63.3$}~\textcolor{teal}{\small{($+1.6$)}} & {$75.0$}~\textcolor{teal}{\small{($+3.3$)}} & {$16.6$}~\textcolor{teal}{\small{($+1.6$)}} & $18.3$~\textcolor{magenta}{\small{($-3.3$)}} & {$15.0$}~\textcolor{teal}{\small{($+3.3$)}} & $11.7$~\small{($-$)} & {$\bf{33.3}$}~\textcolor{teal}{\small{($+1.1$)}} \\
     \midrule
     \midrule
     \multicolumn{8}{c}{(\textit{Zero-Shot-CoT})}\\
    \texttt{gpt-4}&$93.3$&$90.0$&$55.0$&$50.0$&$36.7$&$41.7$&$61.1$\\
     \ +S2L w/ model&$95.0$~\textcolor{teal}{\small{($+1.7$)}}&$91.7\phantom{0}$~\textcolor{teal}{\small{($+1.7$)}}&$38.3$~\textcolor{magenta}{\small{($-16.7$)}}&$46.7$~\textcolor{magenta}{\small{($-3.3$)}}&{$43.3$}~\textcolor{teal}{\small{($+6.6$)}}&{$48.3$}~\textcolor{teal}{\small{($+6.6$)}}&{$60.5$}~\textcolor{magenta}{\small{($-0.6$)}}\\
     \ +S2L w/ rule&{$96.7$}~\textcolor{teal}{\small{($+3.4$)}}&{$100.0$}~\textcolor{teal}{\small{($+10.0$)}}&{$86.7$}~\textcolor{teal}{\small{($+31.7$)}}&{$88.3$}~\textcolor{teal}{\small{($+38.3$)}}&{$50.0$}~\textcolor{teal}{\small{($+13.3$)}}&{$66.7$}~\textcolor{teal}{\small{($+25.0$)}}&{$\bf{81.4}$}~\textcolor{teal}{\small{($+20.3$)}}\\
     \midrule
     \texttt{gpt-3.5-turbo} &$76.7$&$78.3$&$15.0$&$25.0$&$26.7$&$25.0$&$41.1$\\
     \ +S2L w/ model&{$75.0$}~\textcolor{magenta}{\small{($-1.7$)}}&{$71.7$}~\textcolor{magenta}{\small{($-6.6$)}}&{$30.0$}~\textcolor{teal}{\small{($+15.0$)}}&{$35.0$}~\textcolor{teal}{\small{($+10.0$)}}&{$35.0$}~\textcolor{teal}{\small{($+8.3$)}}&{$30.0$}~\textcolor{teal}{\small{($+5.0$)}}&{$\bf{46.1}$}~\textcolor{teal}{\small{($+5.0$)}}\\
     \ +S2L w/ rule&{$75.0$}~\textcolor{magenta}{\small{($-1.7$)}}&{$86.7$}~\textcolor{teal}{\small{($+8.4$)}}&{$30.0$}~\textcolor{teal}{\small{($+15.0$)}}&{$36.7$}~\textcolor{teal}{\small{($+11.7$)}}&{$21.7$}~\textcolor{magenta}{\small{($-5.0$)}}&$25.0$~{\small{($-$)}}&{$\bf{45.8}$}~\textcolor{teal}{\small{($+4.7$)}}\\
     \midrule
     \texttt{openchat-3.5-7b} & $61.7$ & $71.7$ & $15.0$ & $21.6$ & $11.7$ & $11.7$ & $32.2$ \\
     \ +S2L w/ model & {$68.3$}~\textcolor{teal}{\small{($+6.6$)}} & {$78.3$}~\textcolor{teal}{\small{($+6.6$)}} & {$23.3$}~\textcolor{teal}{\small{($+8.3$)}} & {$25.0$}~\textcolor{teal}{\small{($+3.4$)}} & {$25.0$}~\textcolor{teal}{\small{($+13.3$)}} & {$21.7$}~\textcolor{teal}{\small{($+10.0$)}} & {$\bf{40.3}$}~\textcolor{teal}{\small{($+8.1$)}} \\
     \ +S2L w/ rule & {$63.3$}~\textcolor{teal}{\small{($+1.6$)}} & {$75.0$}~\textcolor{teal}{\small{($+3.3$)}} & {$16.6$}~\textcolor{teal}{\small{($+1.6$)}} & $18.3$~\textcolor{magenta}{\small{($-3.3$)}} & {$15.0$}~\textcolor{teal}{\small{($+3.3$)}} & $11.7$~\small{($-$)} & {$\bf{33.3}$}~\textcolor{teal}{\small{($+1.1$)}} \\
     \bottomrule
\end{tabular}
}
\caption{Results for three subtasks on 1D-ARC. $n$ indicates the number of given input-output pairs for finding the patterns. The \textcolor{teal}{values in parentheses} indicate the difference between the baseline and our results. w/ model denotes conversion with LLMs, and w/ rule denotes conversion with manually designed rules using codes.}
\label{table:abstract_reasoning_new}
\end{table*}

\subsection{Abstract Reasoning}
\label{sec:arc}


Abstract reasoning~\citep{webb2023emergent,gendron2023large,wang2023hypothesis} is a type of task that involves summarizing patterns from limited observations. We conduct experiments on subtasks from the 1D-ARC benchmark proposed by~\cite{xu2023llms}, which is a simplified version of the abstract reasoning corpus proposed by \citet{chollet2019measure}.
The 1D-ARC comprises various 1D object-based visual problems, as depicted in Figure~\ref{figure:1d-arc} (a).
To enable LLMs to process these problems, visual information is transformed into a symbol-based representation with sequences of numbers, as illustrated in Figure~\ref{figure:1d-arc} (b).

\begin{figure*}[t!]
	\centering
	\includegraphics[scale=0.96]{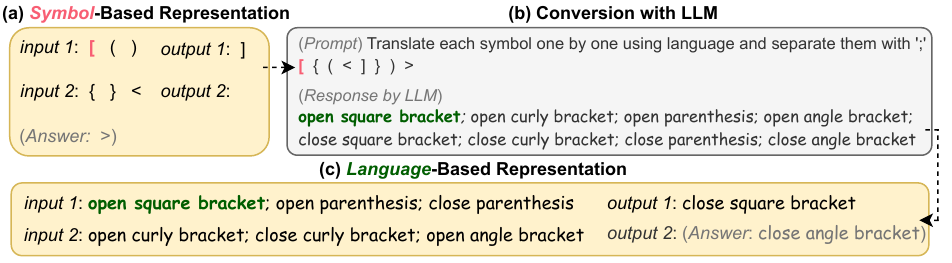}
	\caption{Example of applying symbol-to-language for Dyck language task. We convert every symbol (\textit{e.g.}, ``\texttt{[}'') to its textual description (\textit{e.g.}, ``open square bracket'') by prompting LLMs, and then transform the symbolized problem into language-based representation for both the question and ground truth.}
	\label{figure:dyck}
\end{figure*}

\begin{table*}[t!]
\scalebox{0.93}{
\begin{tabular}{lllllll}
	    \toprule
        \multirow{2.5}*{\textbf{Model}}&\multicolumn{5}{c}{\textbf{Dyck Language}}&\multicolumn{1}{c}{\multirow{2.5}*{\textsc{\textbf{Avg}.}}}\\
        \cmidrule(lr){2-6}
        &$n=1$&$n=2$&$n=3$&$n=4$&$n=5$&\\
    	\midrule
     \multicolumn{7}{c}{(\textit{Zero-Shot})}\\
    \texttt{gpt-4}&$60.0$&$82.2$&$86.6$&$91.4$&$92.2$&$82.5$\\
     \ +S2L w/ model&{$77.6$}~\textcolor{teal}{\small{($+17.6$)}}&{$90.0$}~\textcolor{teal}{\small{($+7.8$)}}&{$95.6$}~\textcolor{teal}{\small{($+9.0$)}}&{$98.2$}~\textcolor{teal}{\small{($+6.8$)}}&{$98.6$}~\textcolor{teal}{\small{($+6.4$)}}&{$\bf{92.0}$}~\textcolor{teal}{\small{($+9.5$)}}\\
     \midrule
     \texttt{gpt-3.5-turbo}&$65.0$&$77.0$&$78.0$&$80.8$&$78.2$&$75.8$\\
     \ +S2L w/ model&{$69.6$}~\textcolor{teal}{\small{($+4.6$)}}&{$82.8$}~\textcolor{teal}{\small{($+5.8$)}}&{$88.2$}~\textcolor{teal}{\small{($+10.2$)}}&{$93.2$}~\textcolor{teal}{\small{($+12.4$)}}&{$94.0$}~\textcolor{teal}{\small{($+15.8$)}}&{$\bf{85.6}$}~\textcolor{teal}{\small{($+9.8$)}}\\
     \midrule
     \texttt{openchat-3.5-7b}&$8.6$&$3.6$&$4.2$&$2.2$&$7.8$&$5.3$\\
     \ +S2L w/ model&{$21.6$}~\textcolor{teal}{\small{($+13.0$)}}&{$22.0$}~\textcolor{teal}{\small{($+18.4$)}}&{$32.0$}~\textcolor{teal}{\small{($+27.8$)}}&{$43.8$}~\textcolor{teal}{\small{($+41.6$)}}&{$59.0$}~\textcolor{teal}{\small{($+51.2$)}}&{$\bf{35.7}$}~\textcolor{teal}{\small{($+30.4$)}}\\
     \midrule
     \midrule
    \multicolumn{7}{c}{(\textit{Zero-Shot-CoT})}\\
    \texttt{gpt-4}&$56.0$&$80.2$&$86.4$&$91.6$&$92.6$&$81.4$\\
     \ +S2L w/ model&{$75.4$}~\textcolor{teal}{\small{($+19.4$)}}&{$91.2$}~\textcolor{teal}{\small{($+11.0$)}}&{$95.8$}~\textcolor{teal}{\small{($+9.4$)}}&{$97.6$}~\textcolor{teal}{\small{($+6.0$)}}&{$98.4$}~\textcolor{teal}{\small{($+5.8$)}}&{$\bf{91.7}$}~\textcolor{teal}{\small{($+10.3$)}}\\
     \midrule
     \texttt{gpt-3.5-turbo}&$73.0$&$75.0$&$74.2$&$80.4$&$78.0$&$76.1$\\
     \ +S2L w/ model&{$68.8$}~\textcolor{magenta}{\small{($-4.2$)}}&{$83.0$}~\textcolor{teal}{\small{($+8.0$)}}&{$90.0$}~\textcolor{teal}{\small{($+15.8$)}}&{$93.6$}~\textcolor{teal}{\small{($+13.2$)}}&{$93.8$}~\textcolor{teal}{\small{($+15.8$)}}&{$\bf{85.8}$}~\textcolor{teal}{\small{($+9.7$)}}\\
     \midrule
     \texttt{openchat-3.5-7b}&$8.6$&$3.6$&$4.2$&$2.2$&$7.8$&$5.3$\\
     \ +S2L w/ model&{$21.6$}~\textcolor{teal}{\small{($+13.0$)}}&{$22.0$}~\textcolor{teal}{\small{($+18.4$)}}&{$32.0$}~\textcolor{teal}{\small{($+27.8$)}}&{$43.8$}~\textcolor{teal}{\small{($+41.6$)}}&{$59.0$}~\textcolor{teal}{\small{($+51.2$)}}&{$\bf{35.7}$}~\textcolor{teal}{\small{($+30.4$)}}\\
     \bottomrule
\end{tabular}
}
\caption{Results for Dyck language task.  $n$ indicates the number of given input-output pairs for finding the patterns. w/ model denotes conversion with LLMs.}
\label{table:result_dyck_language}
\end{table*}

\textbf{Symbol-to-Language.} 
As for the conversion methods, we apply both LLMs and rule-based codes to transform the sequence of numbers into natural language descriptions.
We find that LLMs describe the sequence similarly to humans coincidentally, which employs \textit{merging} or \textit{counting} when describing number sequences. Thus, for the conversion with rules, we implement code as a rule for merging and counting numbers in a sequence.
The specific prompts for LLMs and the rule-based codes are presented in Figure~\ref{figure:1d-arc} (c.1) and Figure~\ref{figure:1d-arc} (d.1).
Due to the potential loss of information in the generated description $l_i^{\rm LLM}$, we attach the language-based representation to each original sequence of numbers to generate answers, as shown in Figure~\ref{figure:1d-arc} (c.2).
On the contrary, the outputs by rule-based codes $l^{\rm rule}_i$ are equal to the original sequence of numbers, thus we directly replace them to get responses, as shown in Figure~\ref{figure:1d-arc} (d.2).

\textbf{Settings and Results.}
We use the Move-1p, Move-2p, and Move-3p tasks (move 1, 2, and 3 pixels forward, respectively) from 1D-ARC, where each task contains 50 problems that involve fixed $n=3$ input-output pairs. We collect and combine the given input-output pairs for each task, resulting in 60 problems for each task with $n=3$ or $n=4$ input-output pairs. The experimental results are presented in Table~\ref{table:abstract_reasoning_new}.
GPT-4 achieves an accuracy above $90.0\%$ on the Move-1p task.
However, the performance drops rapidly to $30$$\sim$$50\%$ in the Move-2p and Move-3p tasks, demonstrating that the model struggles to reason on sequences with slightly more complex patterns. 
This phenomenon is even more evident in ChatGPT and OpenChat, where the overall accuracy is much lower.
Upon employing conversion with LLMs (\textit{i.e.}, S2L w/ model), the results improve by $2.5$$\sim$$8.6\%$, suggesting the positive impact of additional language-based information.
By using conversion with rule-based codes, GPT-4 gets $100\%$ accuracy given $n=4$ input-output pairs, and the performance on Move-2p and Move-3p improves substantially, with $96.7\%$ and $60.0\%$ accuracy, respectively.
Moreover, we observe varying degrees of improvement across models, indicating their differing abilities to understand the additional language-based representation.

\subsection{Dyck Language}
\label{sec:dyck}

Dyck language\footnote{\url{https://raw.githubusercontent.com/google/BIG-bench/main/bigbench/benchmark_tasks/dyck_languages/task.json}.} is a subtask in BigBench~\citep{srivastava2022beyond}, aiming to predict the closing parentheses of a given sequence. To evaluate the inductive reasoning ability over \textit{symbols}, we do not prompt LLMs to ``complete parentheses'' (\textit{i.e.}, prompting LLMs to output the remaining parentheses). Instead, following the settings of the ARC benchmark, we \textit{only} give $n$ input-output pairs and let LLMs to \textit{deduce} the output according to the patterns, as shown in Figure~\ref{figure:dyck} (a).

\begin{figure*}[t!]
	\centering
	\includegraphics[scale=1.0]{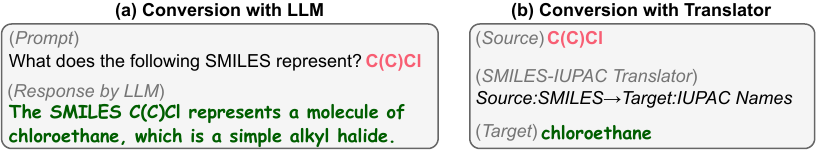}
	\caption{Example of applying symbol-to-language for property prediction. We convert each SMILES to its textual representation by prompting LLMs or using a translator.}
	\label{figure:chem}
\end{figure*}

\begin{table}[t!]
\scalebox{0.843}{
\begin{tabular}{llllllll}
	\toprule
         \multirow{2.5}*{\textbf{Model}}&\multicolumn{3}{c}{\textit{Zero-Shot}}&\multicolumn{3}{c}{\textit{Zero-Shot-CoT}}&\multicolumn{1}{c}{\multirow{2.5}*{\textsc{\textbf{Avg}.}}}\\
         \cmidrule(lr){2-4}\cmidrule(lr){5-7}
        \multirow{2.5}*&\textbf{BACE}&\textbf{BBBP}&\textbf{Tox21}&\textbf{BACE}&\textbf{BBBP}&\textbf{Tox21}&\\
    	\midrule
     \texttt{gpt-4}&$48.2$&$40.2$&$47.2$&$50.4$&$36.8$&$37.4$&$43.4$\\
     \ +S2L w/ model&{$53.0$}~\textcolor{teal}{\small{($+5.2$)}}&{$53.0$}~\textcolor{teal}{\small{($+12.8$)}}&{$48.6$}~\textcolor{teal}{\small{($+1.4$)}}&{$55.6$}~\textcolor{teal}{\small{($+5.2$)}}&{$54.2$}~\textcolor{teal}{\small{($+17.4$)}}&{$44.0$}~\textcolor{teal}{\small{($+6.6$)}}&{$\bf{51.4}$}~\textcolor{teal}{\small{($+8.0$)}}\\
     \ +S2L w/ translator&{$48.4$}~\textcolor{teal}{\small{($+0.2$)}}&{$58.8$}~\textcolor{teal}{\small{($+18.6$)}}&{$49.0$}~\textcolor{teal}{\small{($+1.8$)}}&{$51.2$}~\textcolor{teal}{\small{($+0.8$)}}&{$64.0$}~\textcolor{teal}{\small{($+27.2$)}}&{$45.2$}~\textcolor{teal}{\small{($+7.8$)}}&{$\bf{52.8}$}~\textcolor{teal}{\small{($+9.4$)}}\\
     \midrule
     \texttt{gpt-3.5-turbo}&$52.4$&$24.6$&$34.2$&$44.2$&$32.0$&$35.8$&$37.2$\\
     \ +S2L w/ model&{$53.0$}~\textcolor{teal}{\small{($+0.6$)}}&{$35.0$}~\textcolor{teal}{\small{($+10.4$)}}&{$34.8$}~\textcolor{teal}{\small{($+0.6$)}}&{$48.2$}~\textcolor{teal}{\small{($+4.0$)}}&{$35.8$}~\textcolor{teal}{\small{($+3.8$)}}&{$38.2$}~\textcolor{teal}{\small{($+2.4$)}}&{$\bf{40.8}$}~\textcolor{teal}{\small{($+3.6$)}}\\
      \ +S2L w/ translator&{$54.8$}~\textcolor{teal}{\small{($+2.4$)}}&{$53.8$}~\textcolor{teal}{\small{($+29.2$)}}&{$51.0$}~\textcolor{teal}{\small{($+16.8$)}}&{$48.6$}~\textcolor{teal}{\small{($+4.4$)}}&{$41.4$}~\textcolor{teal}{\small{($+9.4$)}}&{$41.8$}~\textcolor{teal}{\small{($+6.0$)}} & {$\bf{48.6}$}~\textcolor{teal}{\small{($+11.4$)}} \\
      \midrule
     \texttt{openchat-3.5-7b} & $48.2$ & $46.8$ & $62.6$ & $49.6$ & $46.2$ & $62.2$ & $52.6$ \\
     \ +S2L w/ model& {$48.8$}~\textcolor{teal}{\small{($+0.6$)}} & {$56.2$}~\textcolor{teal}{\small{($+9.4$)}} & {$65.0$}~\textcolor{teal}{\small{($+2.4$)}} & $47.8$~\textcolor{magenta}{\small{($-1.8$)}} & {$48.4$}~\textcolor{teal}{\small{($+2.2$)}} & $60.4$~\textcolor{magenta}{\small{($-1.8$)}} & $\bf{54.4}$~\textcolor{teal}{\small{($+1.8$)}}\\
     \ +S2L w/ translator & {$55.8$}~\textcolor{teal}{\small{($+7.6$)}} & {$59.2$}~\textcolor{teal}{\small{($+12.4$)}} & {$67.8$}~\textcolor{teal}{\small{($+5.2$)}} & $52.2$~\textcolor{teal}{\small{($+2.6$)}}  & {$65.2$}~\textcolor{teal}{\small{($+19.0$)}} & $61.4$~\textcolor{magenta}{\small{($-0.8$)}} & $\bf{60.3}$~\textcolor{teal}{\small{($+7.7$)}}\\
    
     \bottomrule
\end{tabular}
}
\caption{Results for three property prediction tasks on ChemLLMBench by using zero-shot inference, zero-shot-CoT inference, and our symbol-to-language.  w/ model denotes conversion with LLMs, and w/ translator denotes conversion with an external translator.}
\label{table:result_chemBenchmark}
\end{table}

\textbf{Symbol-to-Language.} 
The symbols in this task include totally eight different brackets (``\texttt{[]\{\}()<>}''). We consider converting each symbol $s_i$ to natural language description $l_i^{\rm LLM}$ via prompting. Thus we can translate the problem in language-based representations, as shown in Figure~\ref{figure:dyck} (c).

\textit{Remark.} The responses during S2L conversion may not be the same among different LLMs. An interesting phenomenon is that GPT-4 recognizes the symbols ``\texttt{<}'' and ``\texttt{>}'' as ``less than'' and ``greater than'', instead of ``open angle bracket'' and ``close angle bracket''. However, we do \textit{not} consider this to be a mistake and do \textit{not} intend to modify the prompts to ``correct'' the results (\textit{e.g.}, provide some cues indicating these are different types of parentheses). In contrast, we let the LLMs convert the symbols according to their understanding and deduce the final answer through the generated representations by themselves.

\textbf{Settings and Results.} We first set the number of examples as $n=5$ and randomly choose six input-output pairs (five of them as the examples and the rest as the target) from the entire dataset, we repeat 500 times and evaluate the overall accuracy among these 500 questions. Then we gradually remove the last input-output pair to test the ability with fewer examples (\textit{i.e.}, $n=4,3,2,1$).

The results are shown in Table~\ref{table:result_dyck_language}. For GPT-4 and ChatGPT, the performance ranges from $60.0$$\sim$$92.2\%$ and $65.0$$\sim$$78.2\%$, respectively. The accuracy further improves with $+9.5\%$ and $+9.8\%$ by using S2L. For OpenChat, the performance is extremely low with below $10\%$ accuracy and S2L improves the performance by a large margin with $+30.4\%$ on average.

\subsection{Property Prediction}
\label{sec:chem}

We use ChemLLMBench~\citep{guo2023gpt} to predict the chemical property given a molecule's SMILES (simplified molecular-input line-entry system) string. Three datasets are used, including BACE (bindings results for a set of inhibitors of human beta-secretase), BBBP (penetration/non-penetration to the brain-blood barrier), and Tox21 (toxicity of compounds)\footnote{We use the data and prompts from \url{https://github.com/ChemFoundationModels/ChemLLMBench/tree/main/data/property_prediction}. The HIV and ClinTox datasets are excluded due to the biased label distribution.}.

\begin{figure*}[t!]
	\centering
	\includegraphics[scale=0.95]{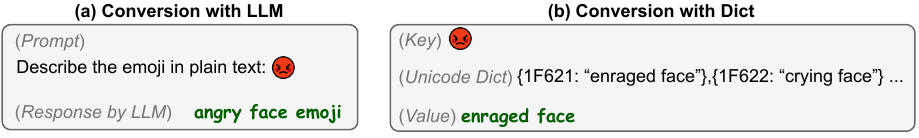}
	\caption{Example of applying symbol-to-language for emotional reranking of emojis. We convert each emoji to its language-based representation by prompting LLMs or using the names from the Unicode dictionary.}
	\label{figure:emoji}
\end{figure*}
\begin{table}
\scalebox{0.78}{
	\begin{tabular}{lccccccccl}
	    \toprule
        \multirow{2.5}*{\textbf{Model}}&\multicolumn{8}{c}{\textbf{EmoTag1200} (Pearson correlation $r$)}&\multicolumn{1}{c}{\multirow{2.5}*{\textsc{\textbf{Avg}.}}}\\
        \cmidrule(lr){2-9}
&\textsc{Anger}&\textsc{Anticipate}&\textsc{Disgust}&\textsc{Fear}&\textsc{Joy}&\textsc{Sadness}&\textsc{Surprise}&\textsc{Trust}&\\
    \midrule
     \multicolumn{10}{c}{(\textit{Zero-Shot})}\\

    \texttt{gpt-4}& $0.855 $&$0.290$&$ 0.782 $&$ 0.809 $&$ 0.887 $&$ 0.903 $&$ 0.531 $&$ 0.731 $&$ 0.724$\\
     \ +S2L w/ model&${0.878} $&$ {0.308} $&$ {0.792} $&$ {0.819} $&$ {0.897} $&$ {0.922} $&$ {0.562} $&$ {0.766} $&$ \bf{0.743}$~\textcolor{teal}{($+0.019$)} \\
     \ +S2L w/ dict&$ {0.856} $&$ {0.345} $&$ 0.753 $&$ {0.813} $&$ {0.894} $&$ {0.926} $&$ {0.616} $&$ {0.760} $&$ \bf{0.745}$~\textcolor{teal}{($+0.021$)}\\
     \midrule
     \texttt{gpt-3.5-turbo}&$ 0.710 $&$0.214$&$ 0.334 $&$ 0.589 $&$ 0.695 $&$0.790 $&$ 0.192 $&$0.560$&$0.510$\\
     \ +S2L w/ model&$0.700$ & $0.156$ & $0.401$ & $0.629$ & $0.786$ & $0.850$& $0.483$ & $0.630$ & $\bf{0.578}$~\textcolor{teal}{($+0.068$)} \\
     \ +S2L w/ dict& $0.757$ & $0.269$ & $0.487$ & $0.696$ & $0.796$ &$0.854$ & $0.285$ & $0.700$& $\bf{0.605}$~\textcolor{teal}{($+0.095$)}\\
     \midrule
     \texttt{openchat-3.5-7b} & $0.413$ & $0.030$ & $0.438$ & $0.264$ & $0.161$ & $0.232$ & $0.221$ & $-0.086\phantom{-}$ & $0.209$ \\
     \ +S2L w/ model & ${0.465}$ & ${0.066}$ & ${0.535}$ & ${0.639}$ & ${0.355}$ & ${0.583}$ & $0.010$ & $-0.118\phantom{-}$ & $\bf{0.317}$~\textcolor{teal}{($+0.108$)} \\
     \ +S2L w/ dict& ${0.572}$ & $0.002$ & ${0.587}$ & ${0.648}$ & ${0.377}$ & ${0.587}$ & $0.023$ & ${-0.078\phantom{-}}$ & $\bf{0.339}$~\textcolor{teal}{($+0.130$)} \\
     \midrule
     \midrule
      \multicolumn{10}{c}{(\textit{Zero-Shot-CoT})}\\
     \texttt{gpt-4}&$0.854$ & $0.205$ & $0.723$ & $0.810$ & $0.889$ & $0.917$ & $0.507$ & $0.744$& $0.706$\\
     \ +S2L w/ model&$0.854$ & ${0.330}$ & $0.705$ & ${0.814}$& $0.889$ & ${0.923}$ & ${0.632}$ & $0.741$ & $\bf{0.736}$~\textcolor{teal}{($+0.030$)}\\
     \ +S2L w/ dict& ${0.865}$ & ${0.337}$ & $0.661$ & ${0.825}$& $0.889$ & ${0.930}$ & ${0.627}$ & $0.730$ & $\bf{0.733}$~\textcolor{teal}{($+0.027$)}\\
     \midrule
     \texttt{gpt-3.5-turbo}&$0.559$& $0.013$ & $0.051$ & $0.156$ & $0.264$ & $0.145$ &$-0.084\phantom{-}$ &$0.060$ & $0.146$\\
     \ +S2L w/ model&${0.702}$ &${0.152}$ & ${0.393}$ & ${0.631 }$& ${0.785}$ & ${0.845}$ &${0.485}$ & ${0.633 }$& $\bf{0.578}$~\textcolor{teal}{($+0.432$)} \\
     \ +S2L w/ dict& ${0.734}$ & ${0.206}$ & ${0.505}$& ${0.664}$ & ${0.799}$ & ${0.853}$ & ${0.354}$ & ${0.543}$ &$\bf{0.582}$~\textcolor{teal}{($+0.436$)}\\
     \midrule
         \texttt{openchat-3.5-7b}  & $0.455$ & $-0.014\phantom{-}$ & $0.277$ & $0.348$ & $0.454$ & $0.741$ & $0.163$ & $0.012$ & $0.305$ \\
    \ +S2L w/ model & ${0.632}$ & ${\phantom{-}0.172\phantom{-}}$ & ${0.438}$ & ${0.532}$ & ${0.669}$ & $0.692$ & ${0.344}$ & ${0.098}$ & $\bf{0.447}$~\textcolor{teal}{($+0.142$)} \\
    \ +S2L w/ dict& ${0.564}$ & $-0.050\phantom{-}$ & ${0.381}$ & ${0.613}$ & ${0.684}$ & $0.581$ & ${0.194}$ & ${0.064}$ & $\bf{0.380}$~\textcolor{teal}{($+0.075$)} \\
     \bottomrule
\end{tabular}
}
\caption{Results for emotion analysis of emojis by using zero-shot inference, zero-shot-CoT inference, and our symbol-to-language. The numbers indicate the Pearson correlation coefficient with ratings by humans.  w/ model denotes conversion with LLMs, and w/ rule denotes conversion with the Unicode dictionary.}
\label{table:Table_emoji_Pearson}
\end{table}

\textbf{Symbol-to-Language.}
We use a unified prompt to transfer each SMILES to its language-based representation $l_i^{\rm LLM}$ in all three datasets, as shown in Figure~\ref{figure:chem} (a). Instead of using LLMs, we further propose S2L with STOUT V2.0~\citep{rajan2021stout}, a translator offering the IUPAC (A universally accepted naming scheme established by the International Union of Pure and Applied Chemistry) name $l_i^{\rm translator}$ of a given SMILES, as shown in Figure~\ref{figure:chem} (b). Finally, we append the obtained information to each SMILES notation as language-enhanced input for LLMs.

{\noindent \textbf{Settings and Results.}} Following~\cite{guo2023gpt}, we randomly sample 500 instances from the full test set and report the averaged results over repeated five times, the results are shown in Table~\ref{table:result_chemBenchmark}. The overall zero-shot performance of GPT-4 and ChatGPT is relatively low, showing the difficulty for LLMs to understand the molecular formula and their chemical property. Zero-shot-CoT does not lead to stable improvement, showing that a single prompt ``\textit{Let's think step by step}'' is not helpful for these types of problems. Our method generally improves the performance to varying degrees (except for slight decreases using OpenChat and zero-shot-CoT settings). For example, the improvement is large for the BBBP dataset ($+9.4$$\sim$$29.2\%$ across models) while it becomes relatively low for the BACE dataset ($+0.2$$\sim$$7.6\%$ across models). Overall, the results show that S2L is effective by providing language-based information that aids in chemical property prediction tasks.

\begin{figure*}[t!]
	\centering
	\includegraphics[scale=1.01]{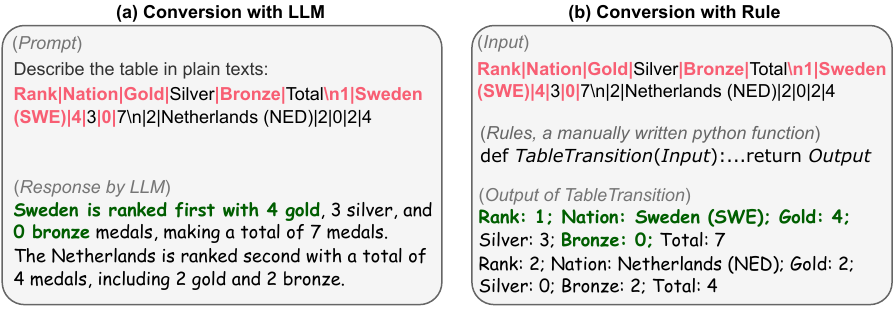}
	\caption{Example of applying symbol-to-language for table question answering task. We convert each table to its textual representation by prompting LLMs or using simple rules implemented in codes.}
	\label{figure:table}
\end{figure*}

\begin{table}
\scalebox{0.875}{
	\begin{tabular}{llllllll}
	    \toprule
        \multirow{3.5}*{\textbf{Model}}&\multicolumn{3}{c}{\textit{Zero-Shot}}&\multicolumn{3}{c}{\textit{Zero-Shot-CoT}}&\multicolumn{1}{c}{\multirow{3.5}*{\textsc{\textbf{Avg}.}}}\\
        \cmidrule(lr){2-4}\cmidrule(lr){5-7}
        &\multicolumn{1}{c}{\textbf{TableQA}}&\multicolumn{1}{c}{\textbf{TableQA}}&\multicolumn{1}{c}{\textbf{TabFact}}&\multicolumn{1}{c}{\textbf{TableQA}}&\multicolumn{1}{c}{\textbf{TableQA}}&\multicolumn{1}{c}{\textbf{TabFact}}&\\   &\multicolumn{1}{c}{{(F1)}}&\multicolumn{1}{c}{{(EM)}}&\multicolumn{1}{c}{{(Acc.)}}&\multicolumn{1}{c}{{(F1)}}&\multicolumn{1}{c}{{(EM)}}&\multicolumn{1}{c}{{(Acc.)}}&\\
    	\midrule
     \texttt{gpt-4}&$82.0$&$79.8$&$93.6$&$80.7$&$76.8$&$93.2$&$84.4$\\
     \ +S2L w/ model&{$84.6$}~\textcolor{teal}{\small{($+2.6$)}}&{$82.0$}~\textcolor{teal}{\small{($+2.2$)}}&{$95.6$}~\textcolor{teal}{\small{($+2.0$)}}&{$82.9$}~\textcolor{teal}{\small{($+2.2$)}}&{$80.2$}~\textcolor{teal}{\small{($+3.4$)}}&{$94.6$}~\textcolor{teal}{\small{($+1.4$)}}&{$\bf{86.7}$}~\textcolor{teal}{\small{($+2.3$)}}\\
          \ +S2L w/ rule&{$86.5$}~\textcolor{teal}{\small{($+4.5$)}}&${84.2}$~\textcolor{teal}{\small{($+4.4$)}}&${95.8}$~\textcolor{teal}{\small{($+2.2$)}}&${84.2}$~\textcolor{teal}{\small{($+3.5$)}}&${80.6}$~\textcolor{teal}{\small{($+3.8$)}}&${96.4}$~\textcolor{teal}{\small{($+3.2$)}}&${\bf{88.0}}$~\textcolor{teal}{\small{($+3.6$)}}\\
          \midrule
     \texttt{gpt-3.5-turbo}&$66.4$&$63.0$&$82.0$&$73.4$&$69.8$&$77.0$&$71.9$\\
     \ +S2L w/ model&{$69.0$}~\textcolor{teal}{\small{($+2.6$)}}&{$66.0$}~\textcolor{teal}{\small{($+3.0$)}}&{$84.6$}~\textcolor{teal}{\small{($+2.6$)}}&{$73.6$}~\textcolor{teal}{\small{($+0.2$)}}&{$71.2$}~\textcolor{teal}{\small{($+1.4$)}}&{$83.0$}~\textcolor{teal}{\small{($+6.0$)}}&{$\bf{74.6}$}~\textcolor{teal}{\small{($+2.7$)}}\\
      \ +S2L w/ rule&{$68.5$}~\textcolor{teal}{\small{($+2.1$)}}&${64.8}$~\textcolor{teal}{\small{($+1.8$)}}&${86.2}$~\textcolor{teal}{\small{($+4.2$)}}&${77.0}$~\textcolor{teal}{\small{($+3.6$)}}&${72.8}$~\textcolor{teal}{\small{($+3.0$)}}&${83.0}$~\textcolor{teal}{\small{($+6.0$)}}&${\bf{75.4}}$~\textcolor{teal}{\small{($+3.5$)}}\\
      \midrule
     \texttt{openchat-3.5-7b} & $61.7$ & $58.2$ & $79.0$ & $60.8$ & $57.0$ & $83.8$ & $66.8$ \\
      \ +S2L w/ model&{$64.1$}~\textcolor{teal}{\small{($+2.4$)}} & ${59.0}$~\textcolor{teal}{\small{($+0.8$)}} & ${81.0}$~\textcolor{teal}{\small{($+2.0$)}} & ${64.2}$~\textcolor{teal}{\small{($+3.4$)}} & ${60.4}$~\textcolor{teal}{\small{($+3.4$)}} & ${83.2}$~\textcolor{magenta}{\small{($-0.6$)}} & ${\bf{68.6}}$~\textcolor{teal}{\small{($+1.8$)}} \\
      \ +S2L w/ rule& ${62.1}$~\textcolor{teal}{\small{($+0.4$)}} & ${58.8}$~\textcolor{teal}{\small{($+0.6$)}} & ${83.0}$~\textcolor{teal}{\small{($+4.0$)}} & ${66.1}$~\textcolor{teal}{\small{($+5.3$)}} & ${63.0}$~\textcolor{teal}{\small{($+6.0$)}} & ${84.6}$~\textcolor{teal}{\small{($+0.8$)}} & ${\bf{69.6}}$~\textcolor{teal}{\small{($+2.8$)}} \\
     \bottomrule
\end{tabular}
}
\caption{Results for table question answering and table fact verification tasks using zero-shot inference, zero-shot-CoT inference, and our symbol-to-language. w/
model denotes conversion with LLMs, and w/ rule denotes conversion with manually designed rules by codes for aligning contents from tables.}
\label{table:results_table}
\end{table}

\subsection{Emotion Analysis of Emojis}
\label{sec:emoji}

We use the EmoTag1200~\citep{shoeb-de-melo-2020-emotag1200} for analyzing the emotions of the emojis. Specifically, 150 most frequently used emojis are used and the task is to score each of them from $0$$\sim$$1$ based on eight basic emotions, including \textit{anger}, \textit{anticipation}, \textit{disgust}, \textit{fear}, \textit{joy}, \textit{sadness}, \textit{surprise}, and \textit{trust}.

\textbf{Symbol-to-Language.}
To understand emojis with language-based information, we get the description $l_i^{\rm LLM}$ by prompting LLMs or directly obtaining the names $l_i^{\rm dict}$ from the Unicode dictionary, as shown in Figure~\ref{figure:emoji}. 

\textbf{Settings and Results.}
We use the Pearson correlation coefficient between the predictions and ratings by humans for evaluation, and the results are shown in Table~\ref{table:Table_emoji_Pearson}. GPT-4 gives a relatively high correlation coefficient of $0.724$. However, ChatGPT and OpenChat perform badly with only $0.510$ and $0.209$ correlation coefficients, respectively. By using zero-shot-CoT, the performance even drops for GPT-4 and ChatGPT, showing the limitations in LLMs' emoji understanding ability. When using S2L with either model or dictionary, the performance improves to different degrees, showing that the language information can also help understand emoji-based symbols.

\subsection{Table Understanding}
\label{sec:table}

For structured data, we follow~\cite{chen-2023-large} to evaluate the capability of LLMs on table reasoning. Specifically, we use WikiTableQuestions~\citep{pasupat-liang-2015-compositional} for evaluating table-based question answering, which consists of complex questions based on Wikipedia tables. We also use TabFact~\citep{Chen2020TabFact} for fact verification, which consists of claims annotated by the crowd workers based on tables. 

\textbf{Symbol-to-Language.}
As shown in Figure~\ref{figure:table}, S2L describes every table in plain text $l_i^{\rm LLM}$ by prompting. Alternatively, we can get the representation $l_i^{\rm rule}$ by using simple rule-based codes to align the content with the table header according to the delimiters ``\texttt{|}'' row by row. Then we append the external natural language information to the original symbol-based representation for each question.

\textbf{Settings and Results.}
For each task, we evaluated 500 pairs of tables and questions and the results are shown in Table~\ref{table:results_table}. The overall performance for different models is relatively high compared with previous symbol-only tasks, for example, GPT-4 gives around $79.8\%$ exact match score and $93.6\%$ accuracy for question answering and fact verification, respectively. Nevertheless, S2L with model can consistently bring $+1.8$$\sim$$2.3\%$ improvements, showing that external natural language information is effective. We find that S2L with rule further leads to $+2.8$$\sim$$3.6\%$ improvements, indicating that simple cues with aligned information between content and header for each row can already make a positive impact on table understanding.

\begin{figure*}[t!]
	\centering
	\includegraphics[scale=0.98]{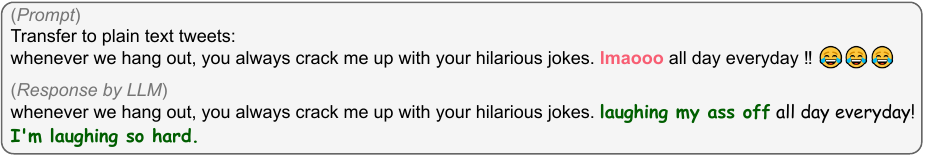}
	\caption{Example of applying symbol-to-language for sentiment classification of tweets. We convert each tweet to plain text by prompting LLMs.}
	\label{figure:tweet}
\end{figure*}

\begin{table}
\scalebox{0.87}{
	\begin{tabular}{llllllll}
	    \toprule
             \multirow{3.5}*{\textbf{Model}}&\multicolumn{3}{c}{\textit{Zero-Shot}}&\multicolumn{3}{c}{\textit{Zero-Shot-CoT}}&\multicolumn{1}{c}{\multirow{3.5}*{\textsc{\textbf{Avg}.}}}\\
             \cmidrule(lr){2-4}\cmidrule(lr){5-7}
        &\multicolumn{1}{c}{\textbf{P-Stance}}&\multicolumn{1}{c}{\textbf{P-Stance}}&\multicolumn{1}{c}{\textbf{Sentiment}}&\multicolumn{1}{c}{\textbf{P-Stance}}&\multicolumn{1}{c}{\textbf{P-Stance}}&\multicolumn{1}{c}{\textbf{Sentiment}}&\\
        
        &\multicolumn{1}{c}{(Acc.)}&\multicolumn{1}{c}{(F1)}&\multicolumn{1}{c}{(Acc.)}&\multicolumn{1}{c}{(Acc.)}&\multicolumn{1}{c}{(F1)}&\multicolumn{1}{c}{(Acc.)}&\\
    	\midrule
     \texttt{gpt-4}&$86.2$&$86.7$&$89.4$&$83.8$&$84.0$&$86.1$& $86.0$\\
     \ +S2L w/ model&${87.1}$~\textcolor{teal}{\small{($+0.9$)}}&${88.2}$~\textcolor{teal}{\small{($+1.5$)}}&${90.3}$~\textcolor{teal}{\small{($+0.9$)}}&${86.6}$~\textcolor{teal}{\small{($+2.8$)}}&${87.5}$~\textcolor{teal}{\small{($+3.5)$}}&$87.2$~\textcolor{teal}{\small{($+1.1)$}}& $\bf{87.8}$~\textcolor{teal}{\small{($+1.8$)}}\\
     \midrule
     \texttt{gpt-3.5-turbo}&$65.3$&$68.6$&$83.7$&$61.5$&$60.8$&$76.5$& $69.4$\\
     \ +S2L w/ model&${71.0}$~\textcolor{teal}{\small{($+5.7$)}}&${71.5}$~\textcolor{teal}{\small{($+2.9$)}}&${89.9}$~\textcolor{teal}{\small{($+6.2$)}}&${64.7}$~\textcolor{teal}{\small{($+3.2$)}}&${63.5}$~\textcolor{teal}{\small{($+2.7$)}}&${78.4}$~\textcolor{teal}{\small{($+1.9$)}}& $\bf{73.2}$~\textcolor{teal}{\small{($+3.8$)}}\\
     \midrule
     \texttt{openchat-3.5-7b} & $70.9$ & $66.7$ & $89.1$ & $72.2$ & $69.1$ & $84.8$ & $75.5$\\
      \ +S2L w/ model&${71.4}$~\textcolor{teal}{\small{($+0.5$)}} & ${67.5}$~\textcolor{teal}{\small{($+0.8$)}} & ${89.0}$~\textcolor{magenta}{\small{($-0.1$)}} & ${77.1}$~\textcolor{teal}{\small{($+4.9$)}} & ${75.8}$~\textcolor{teal}{\small{($+6.7$)}} & {$82.1$}~\textcolor{magenta}{\small{($-2.7$)}} & $\bf{77.2}$~\textcolor{teal}{\small{($+1.7$)}}\\
     \bottomrule
\end{tabular}
}
\caption{Results for stance detection and sentiment classification in social media using zero-shot inference, zero-shot-CoT inference, and our symbol-to-language.}
\label{table:results_tweet}
\end{table}

\subsection{Tweet Analysis}
\label{sec:tweet}
We analyze the text in social media and use the TweetSentimentExtraction dataset from Massive Text Embedding Benchmark~\citep{muennighoff-etal-2023-mteb} for sentiment classification. Additionally, we follow~\cite{zhang2023investigating} by using the P-Stance~\citep{pstance} dataset for stance detection.

\textbf{Symbol-to-Language.}
There is a plethora of non-natural language expressions on Tweet, including abbreviations (\textit{e.g.}, LOL: ``Laughing Out Loud''), slang (\textit{e.g.}, FTW: ``For the Win''), hashtags (\textit{e.g.}, \#Trump), emojis (\textit{e.g.}, \includegraphics[width=2.6mm, height=3.3mm]{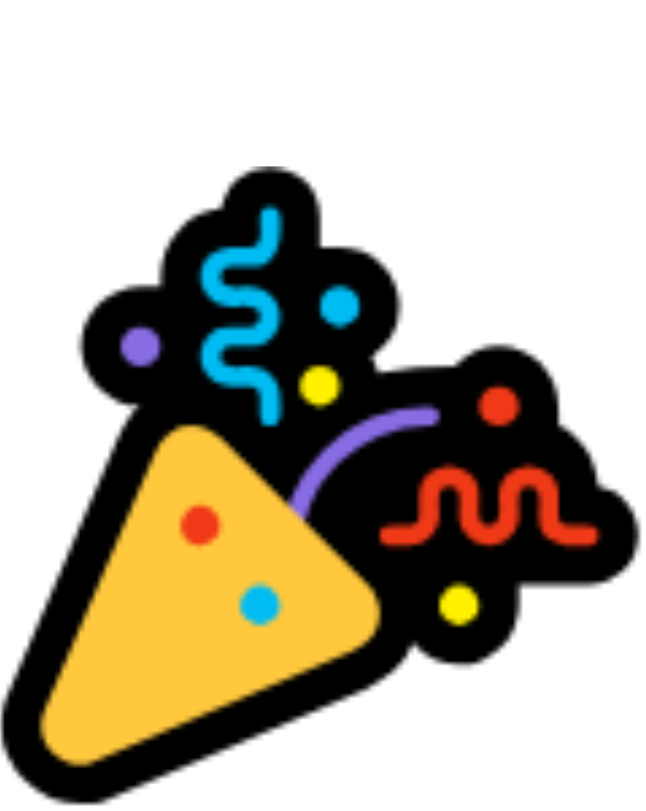}), \textit{etc}. We convert the entire tweet to plain text $l_i^{\rm LLM}$ via prompting, as shown in Figure~\ref{figure:tweet}, and then we use it as external input for each question.

\textbf{Settings and Results.}
For sentiment classification, we predict sentiment polarity from either \textit{positive} or \textit{negative} in a total of 2,104 tweets. For stance detection, we evaluate the attitude towards ``Donald Trump'' from either \textit{favor} or \textit{against} in 777 test tweets. The results are shown in Table~\ref{table:results_tweet}. Similar to other tasks, the zero-shot-CoT method sometimes causes a decrease in accuracy. Specifically, we find that simply adding the prompt ``\textit{Let's think step by step}'' leads to more \textit{neutral} responses, even though there are all texts with 2-way labels (\textit{i.e.}, \textit{positive}/\textit{negative} and \textit{favor}/\textit{against}). In general, our S2L can improve the performance under both zero-shot and zero-shot-CoT settings for GPT-4 and ChatGPT models, except that there is a slight decrease using the OpenChat model for sentiment classification.

\section{Discussions}
From the experimental results, our proposed symbol-to-language shows a significant improvement in tasks such as abstract reasoning, Dyck language, and chemical property prediction. It also achieves slight gains in a range of NLP-related tasks. Additionally, compared to zero-shot-CoT, our approach exhibits more stable and effective improvements in solving symbol-related problems.

We further analyze and discuss both the advantages (\textit{i.e.}, \textit{how does it take effect?}) and limitations (\textit{i.e.}, \textit{in which scenarios does it still not have an impact?}) of the proposed symbol-to-language method.

\textbf{Advantages.} Directly solving symbol-related problems can be difficult for LLMs due to various reasons. We show that S2L can offer some distinct types of language-based information that are important for better solving the mentioned tasks: 

\indent{\textit{Precise Information}.} As shown in the 1D-ARC tasks, the language-based representations by using rules can reflect the information of sequences precisely, which can compensate for the limitations of language models such as counting numbers, thus enhancing their ability to summarize patterns and deduce the results.

\textit{Co-occurrence Information.} S2L conversion offers co-occurrence information between contexts and task-level labels. For example, the descriptions of emojis (\textit{e.g.}, ``angry face'') and emotional dimensions (\textit{e.g.}, ``anger'') for emoji analysis, and the plain text of abbreviations (\textit{e.g.}, ``laughing my ass off'') and sentiment polarities (\textit{e.g.}, ``positive'') for sentiment classification. These language-level co-occurrences can offer complementary information for symbol-based problems.

\textit{Alignment Information.} Language-based representations can also offer aligned information, which is difficult to extract directly from symbol-based representations. For example, the aligned relationship between ``open'' and ``close'' brackets for the Dyck language task, and the alignment between table contents and header. These explicitly aligned contexts can somewhat help LLMs reasoning on complicated symbol-based tasks.

\textbf{Limitations.} Although we verified S2L on different models across various tasks, there are still some limitations. First, not all non-natural language representations can be easily converted into natural language. For example, the original 2D visual problems from the ARC dataset~\citep{chollet2019measure} are still difficult to describe in language-based representations, though \cite{Xu_Khalil_Sanner_2023} try some ``object-based'' representations for a tiny portion of the dataset. Second, for tasks that we can not rely on external tools with sufficient prior knowledge, prompting LLMs may generate incorrect descriptions or explanations of the symbols due to hallucinations, which may mislead the understanding of symbols that were originally comprehensible directly.

\section{Conclusion}
We propose symbol-to-language, a tuning-free method that converts symbol-based to language-based representation for solving a series of symbol-related problems using large language models. Experiments on GPT-4, ChatGPT, and OpenChat across eight tasks show that symbol-to-language can significantly improve the performance on tasks such as abstract reasoning, Dyck language, and chemical property prediction. We hope to further harness the power of language, leverage the advantages of language-based representations, uncover diverse forms of knowledge expressed in natural language, and explore the untapped potential of large language models to play roles in more scenarios.

\bibliography{references}  
\bibliographystyle{iclr_conference}

\appendix
\section{Case Study}
\label{sec:cases}
\begin{figure*}[t!]
\centering
\includegraphics[scale=0.75]{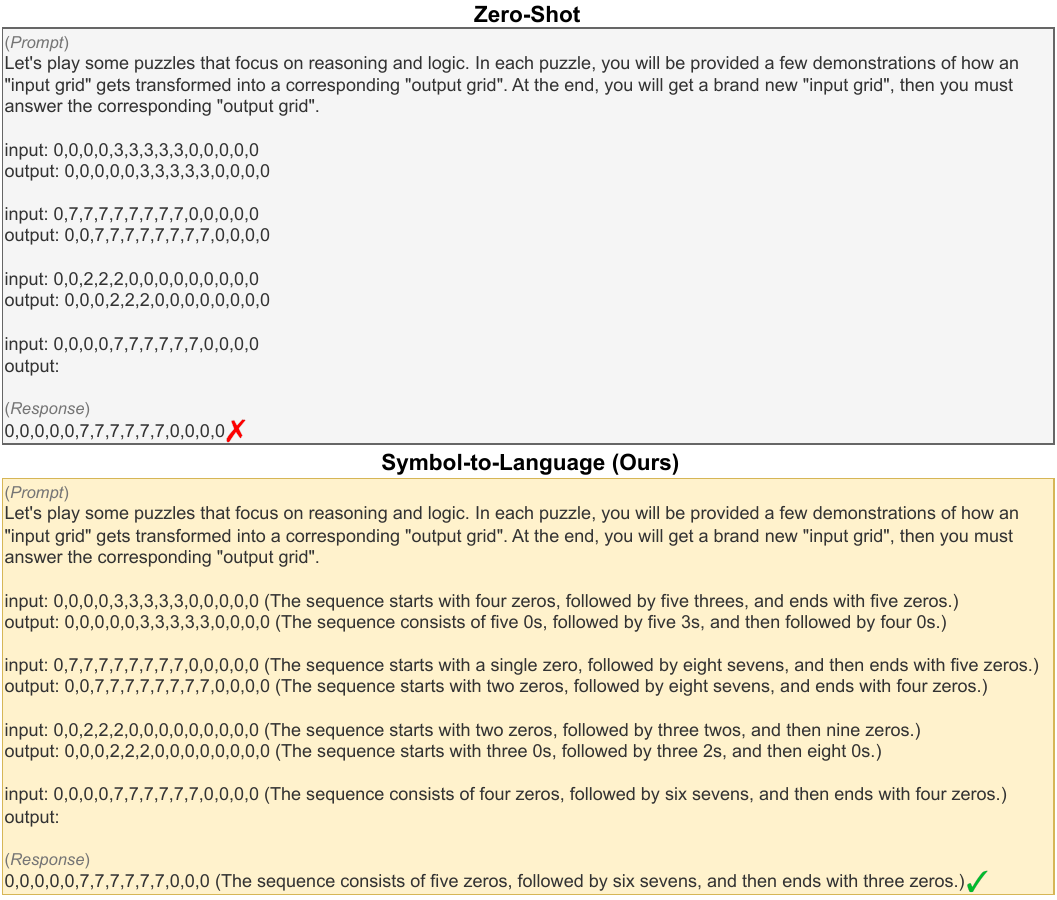}
\caption{Example of response by zero-shot reasoning and our symbol-to-language on 1D-ARC task.}
\label{fig:case1}
\end{figure*}

\begin{figure*}[t!]
\centering
\includegraphics[scale=0.75]{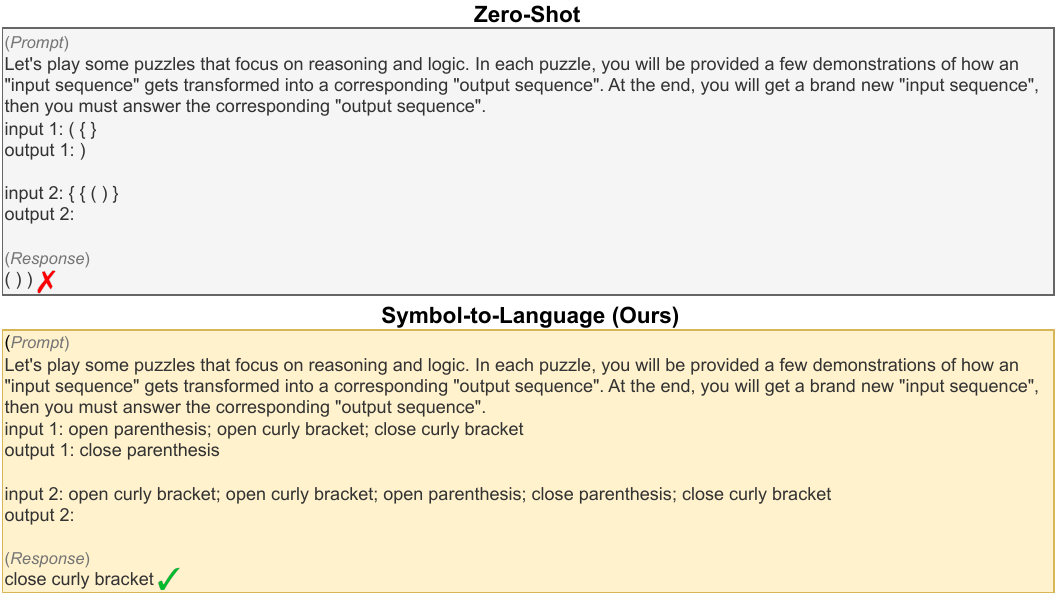}
\caption{Example of response by zero-shot reasoning and our symbol-to-language on Dyck Language.}
\label{fig:case2}
\end{figure*}

\begin{figure*}[t!]
\centering
\includegraphics[scale=0.75]{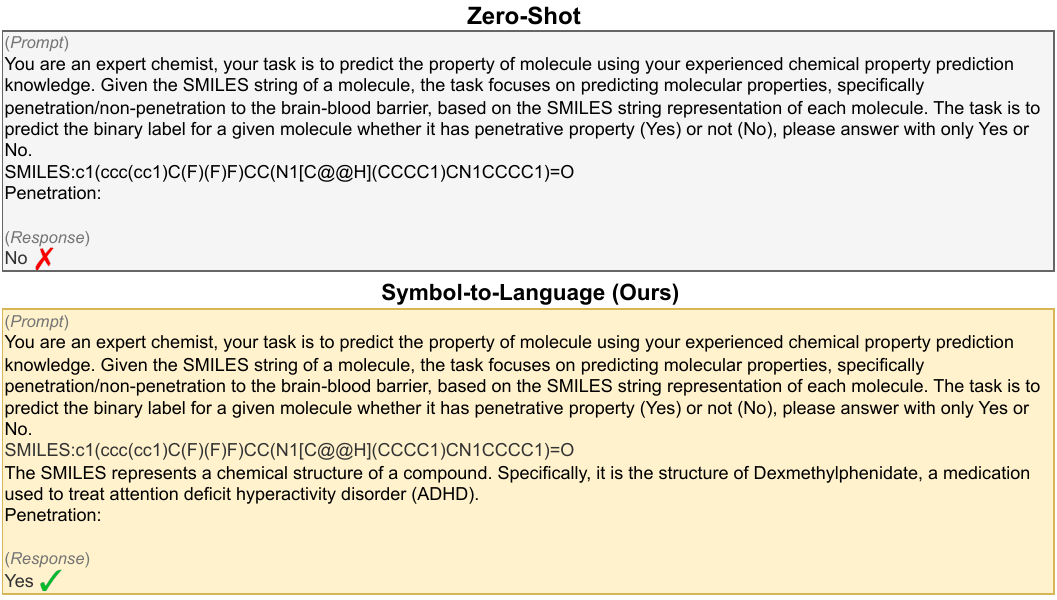}
\caption{Example of response by zero-shot reasoning and our symbol-to-language on property prediction.}
\label{fig:case3}
\end{figure*}

\begin{figure*}[t!]
\centering
\includegraphics[scale=0.75]{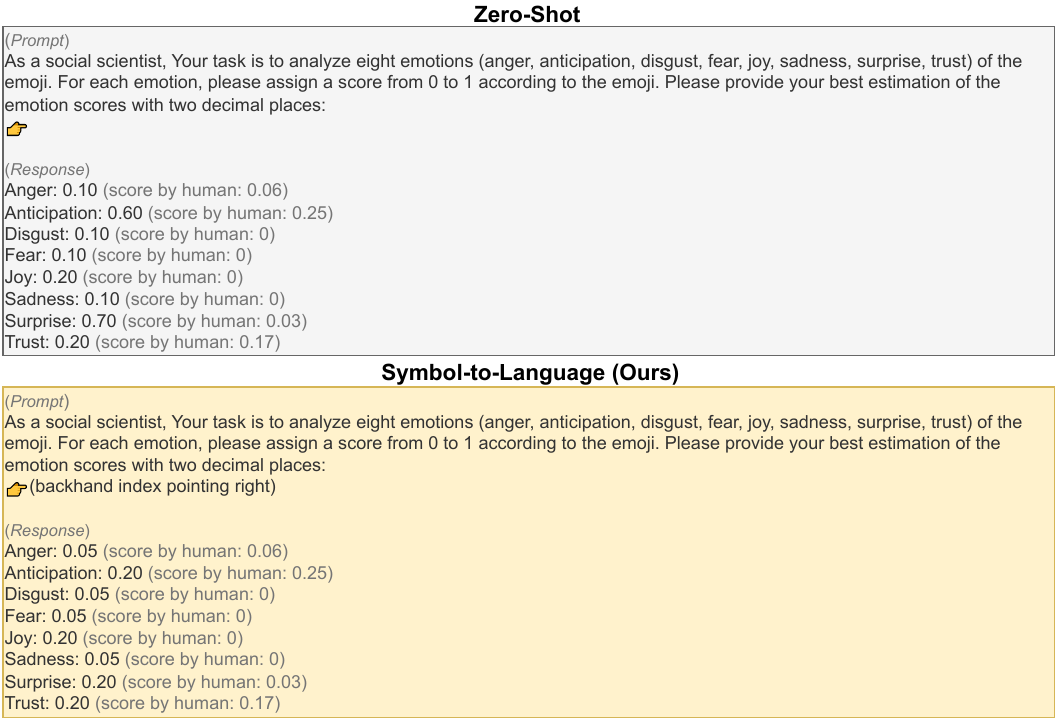}
\caption{Example of response by zero-shot reasoning and our symbol-to-language on emotion analysis of emojis.}
\label{fig:case4}
\end{figure*}

\begin{figure*}[t!]
\centering
\includegraphics[scale=0.75]{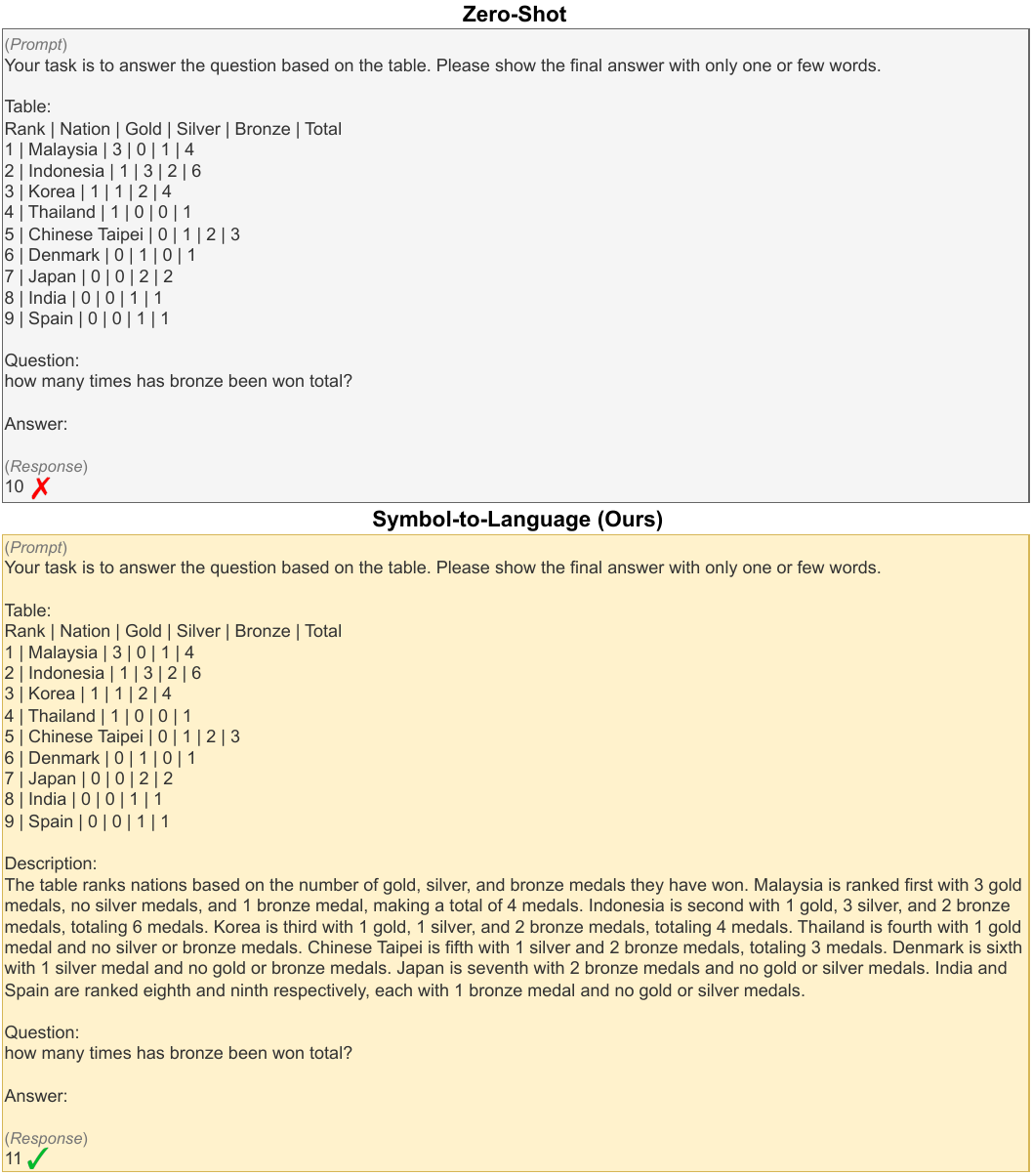}
\caption{Example of response by zero-shot reasoning and our symbol-to-language on table question answering.}
\label{fig:case5}
\end{figure*}

\begin{figure*}[t!]
\centering
\includegraphics[scale=0.75]{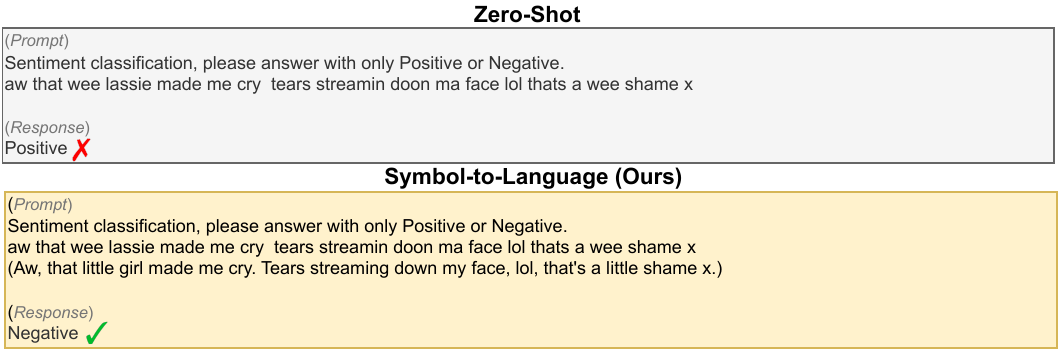}
\caption{Example of response by zero-shot reasoning and our symbol-to-language on sentiment analysis in tweet.}
\label{fig:case7}
\end{figure*}

We show some examples in Figures~\ref{fig:case1},~\ref{fig:case2},~\ref{fig:case3},~\ref{fig:case4},~\ref{fig:case5}, and~\ref{fig:case7} where our symbol-to-language gives correct responses using the GPT-4 model. Please refer to \url{https://github.com/THUNLP-MT/symbol2language} for detailed results.

\end{document}